\documentclass[11pt]{article}
\usepackage{acl2016}
\usepackage{times}
\usepackage{tabularx}
\usepackage{url}
\usepackage{graphicx}
\usepackage{caption}
\usepackage{float}
\usepackage{algorithmicx}
\usepackage[ruled]{algorithm}
\usepackage{algpseudocode}
\usepackage{amssymb,amsmath}
\usepackage{color}
\usepackage{bm}
\usepackage{makecell}
\usepackage{array}

\aclfinalcopy 
\def\aclpaperid{717} 

\usepackage{etoolbox}
\newcommand{\zerodisplayskips}{
  \setlength{\abovedisplayskip}{10pt}
  \setlength{\belowdisplayskip}{10pt}
  \setlength{\abovedisplayshortskip}{4pt}
  \setlength{\belowdisplayshortskip}{4pt}}
\appto{\normalsize}{\zerodisplayskips}
\appto{\small}{\zerodisplayskips}
\appto{\footnotesize}{\zerodisplayskips}

\let\oldenumerate\enumerate
\renewcommand{\enumerate}{
  \vspace{-0.7\topsep} 
  \oldenumerate
  \setlength{\itemsep}{5pt}
  \setlength{\parskip}{0pt}
  \setlength{\parsep}{0pt}
  \setlength{\topsep}{0pt}
  \setlength{\partopsep}{0pt}
}
\let\olditemize\itemize
\renewcommand{\itemize}{
  \vspace{-0.7\topsep}
  \olditemize
  \setlength{\itemsep}{1.5mm}
  \setlength{\parskip}{0pt}
  \setlength{\parsep}{0pt}
  \setlength{\topsep}{0pt}
  \setlength{\partopsep}{0pt}
}

\newcommand*{\Resize}[2]{\resizebox{#1}{!}{$#2$}}
\newcommand{\alns}[1] {
	\begin{align*} #1 \end{align*}
}
\DeclareMathOperator*{\argmax}{argmax}
\def\mathhyphen{{\hbox{-}}}
\newcommand{\bR} { \bm{R} }

\newcommand{\bh} { \bm{h} }

\newcommand{\bW} { \bm{W} }
\newcommand{\bw} { \bm{w} }
\newcommand{\bb} { \bm{b} }

\newcommand{\pir} { \pi^{\text{ref}} }
\newcommand{\scl}{s_c}
\newcommand{\sm}{s_m}
\newcommand{\dm}{d}
\newcommand{\dcl} {d}
\newcommand{\cm} {c_m}
\newcommand{\repm} {\bm{r}_m}
\newcommand{\repcl} {\bm{r}_c}

\newcommand\Tstrut{\rule{0pt}{2.6ex}}         
\newcommand\Bstrut{\rule[-1.0ex]{0pt}{0pt}}   
\newcommand{\mline}{\Xhline{1.5\arrayrulewidth}}
\newcommand{\tline}{\Xhline{2.5\arrayrulewidth}}
\newcommand{\lone}	{\Tstrut \Bstrut \\ \mline}
\newcommand{\ttop}{\tline}
\newcommand{\tbottom}{\Bstrut \\ \tline}
\newcommand{\precaption}{\vspace{-2mm}}

\newcommand{\xhdr}[1]{\vspace{1.7mm}\noindent{{\bf #1.}}}

\title{Improving Coreference Resolution by Learning Entity-Level Distributed Representations}

\author{
	Kevin Clark\\
	Computer Science Department\\
	Stanford University \\
	{\tt kevclark@cs.stanford.edu}
\And
	Christopher D. Manning \\
  	Computer Science Department \\
  	Stanford University \\
  	{\tt manning@cs.stanford.edu} \\}

\begin{document}

\maketitle

\begin{abstract}
{ A long-standing challenge in coreference resolution has been the incorporation of entity-level information -- features defined over clusters of mentions instead of mention pairs. We present a neural network based coreference system that produces high-dimensional vector representations for pairs of coreference clusters. Using these representations, our system learns when combining clusters is desirable. We train the system with a learning-to-search algorithm that teaches it which local decisions (cluster merges) will lead to a high-scoring final coreference partition. The system substantially outperforms the current state-of-the-art on the English and Chinese portions of the  CoNLL 2012 Shared Task dataset despite using few hand-engineered features. }
\end{abstract}

%------------------------------------------------------------------------------------------------

\section{Introduction}
Coreference resolution, the task of identifying which mentions in a text refer to the same real-world entity, is fundamentally a clustering problem. However, many recent state-of-the-art coreference systems operate solely by linking pairs of mentions together \cite{durrett2013easy,martschat2015latent,wiseman2015learning}. 

An alternative approach is to use agglomerative clustering, treating each mention as a singleton cluster at the outset and then repeatedly merging clusters of mentions deemed to be referring to the same entity. Such systems can take advantage of entity-level information, i.e., features between clusters of mentions instead of between just two mentions. As an example for why this is useful, it is clear that the clusters \{{\it Bill Clinton}\} and \{{\it Clinton}, {\it she}\} are not referring to the same entity, but it is ambiguous whether the pair of mentions {\it Bill Clinton} and {\it Clinton} are coreferent. 

Previous work has incorporated entity-level information through features that capture hard constraints like having gender or number agreement between clusters \cite{raghunathan2010multi,durrett2013decentralized}. In this work, we instead train a deep neural network to build distributed representations of pairs of coreference clusters. This captures entity-level information with a large number of learned, continuous features instead of a small number of hand-crafted categorical ones.

Using the cluster-pair representations, our network learns when combining two coreference clusters is desirable. At test time it builds up coreference clusters incrementally, starting with each mention in its own cluster and then merging a pair of clusters each step. It makes these decisions with a novel easy-first cluster-ranking procedure that combines the strengths of cluster-ranking \cite{rahman2011narrowing} and easy-first \cite{stoyanov2012easy} coreference algorithms. 

Training incremental coreference systems is challenging because 
the coreference decisions facing a model depend on previous decisions it has already made. We address this by using a learning-to-search algorithm inspired by SEARN \cite{daume2009search} to train our neural network. This approach allows the model to learn which action (a cluster merge) available from the current state (a partially completed coreference clustering) will eventually lead to a high-scoring coreference partition.

Our system uses little manual feature engineering, which means it is easily extended to multiple languages. We evaluate our system on the English and Chinese portions of the CoNLL 2012 Shared Task dataset. The cluster-ranking model significantly outperforms a mention-ranking model that does not use entity-level information. We also show that using an easy-first strategy improves the performance of the cluster-ranking model. Our final system achieves CoNLL F$_1$ scores of 65.29 for English and 63.66 for Chinese, substantially outperforming other state-of-the-art systems.\footnote{Code and trained models are available at \url{https://github.com/clarkkev/deep-coref.}}

%------------------------------------------------------------------------------------------------

\section{System Architecture}
Our cluster-ranking model is a single neural network that learns which coreference cluster merges are desirable. However, it is helpful to think of the network as being composed of distinct sub-networks. The {\it mention-pair encoder} produces distributed representations for pairs of mentions by passing relevant features through a feedforward neural network. The {\it cluster-pair encoder} produces distributed representations for pairs of clusters by applying a pooling operation over the representations of relevant mention pairs, i.e., pairs where one mention is in each cluster. The {\it cluster-ranking model} then scores pairs of clusters by passing their representations through a single neural network layer.

\begin{figure}[bth]
\begin{center}
\vspace{0mm}
\includegraphics[width=0.4\textwidth]{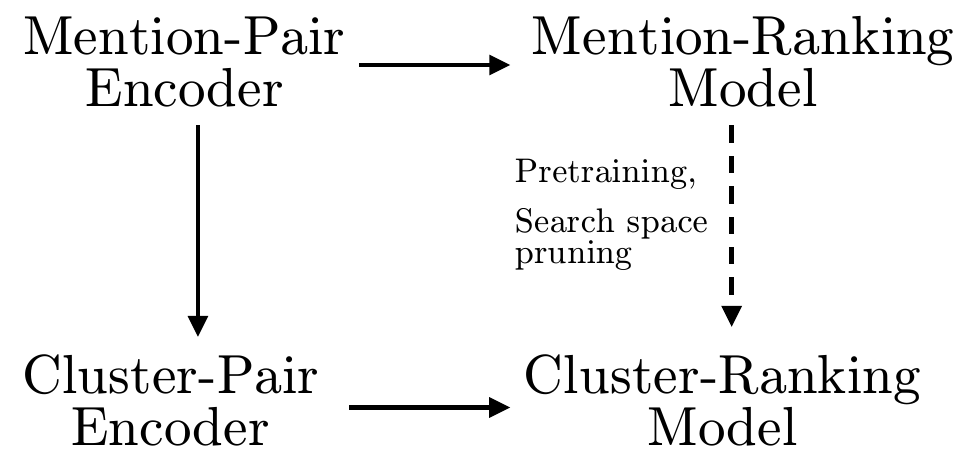}
\vspace{-2mm}
\end{center}
\caption{System architecture. Solid arrows indicate one neural network is used as a component of the other; the dashed arrow indicates other dependencies.}
\vspace{0mm}
\label{fig:architecture}
\end{figure}

We also train a {\it mention-ranking model} that scores pairs of mentions by passing their representations through a single neural network layer. Its parameters are used to initialize the cluster-ranking model, and the scores it produces are used to prune which candidate cluster merges the cluster-ranking model considers, allowing the cluster-ranking model to run much faster. The system architecture is summarized in Figure~\ref{fig:architecture}.

\section{Building Representations}
In this section, we describe the neural networks producing distributed representations of pairs of mentions and pairs of coreference clusters. We assume that a set of mentions has already been extracted from each document using a method such as the one in Raghunathan et al.\ (2010)\nocite{raghunathan2010multi}. 

\subsection{Mention-Pair Encoder}
\begin{figure}[t]
\begin{center}
\includegraphics[width=0.48\textwidth]{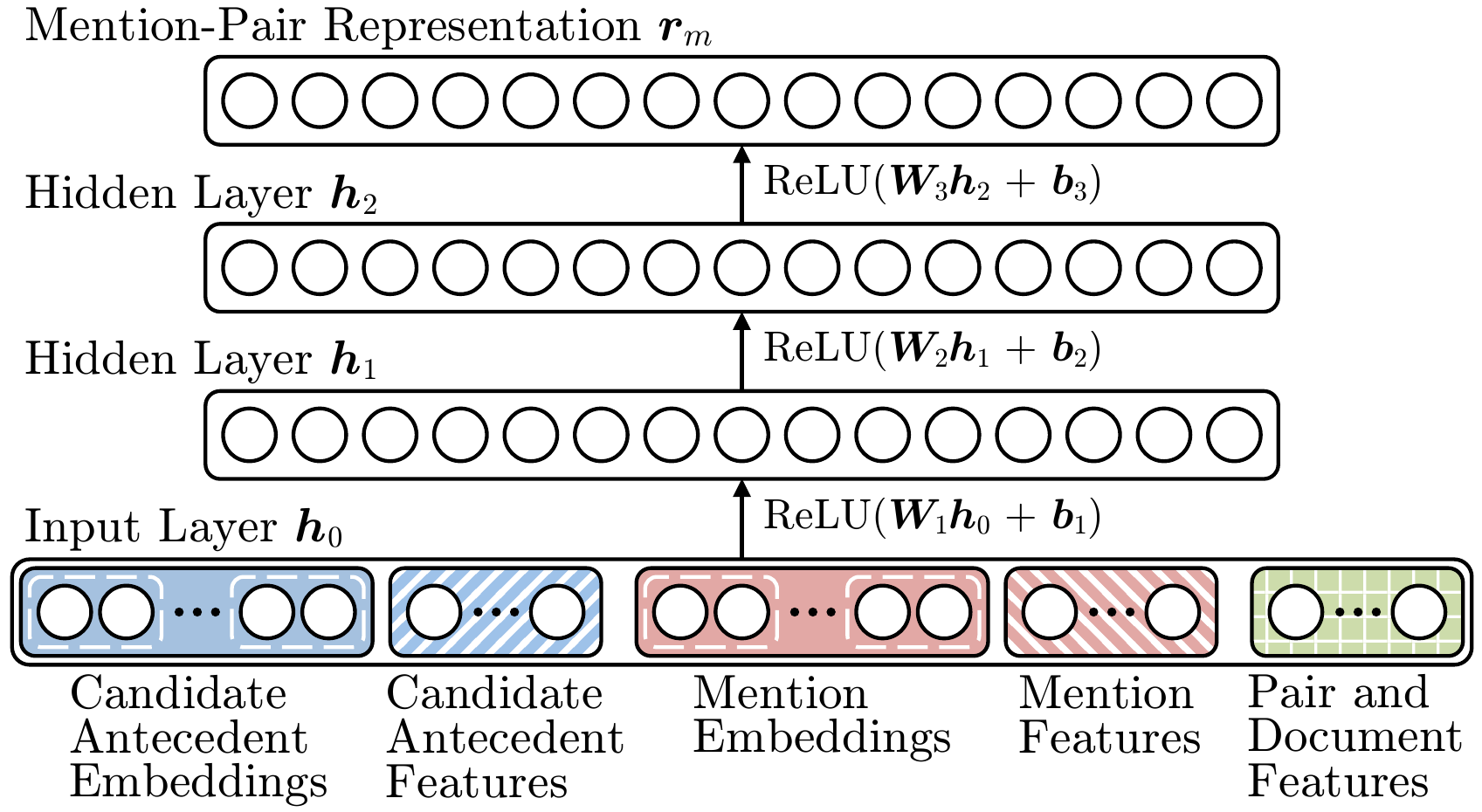}
\vspace{-6mm}
\end{center}
\caption{Mention-pair encoder.}
\label{fig:pair}
\end{figure}

Given a mention $m$ and candidate antecedent $a$, the mention-pair encoder produces a distributed representation of the pair $\repm(a, m) \in \mathbb{R}^{\dm}$ with a feedforward neural network, which is shown in Figure~\ref{fig:pair}. The candidate antecedent may be any mention that occurs before $m$ in the document or {\sc na}, indicating that $m$ has no antecedent. We also experimented with models based on Long Short-Term Memory recurrent neural networks \cite{hochreiter1997long}, but found these to perform slightly worse when used in an end-to-end coreference system due to heavy overfitting to the training data. \\

\xhdr{Input Layer}
For each mention, the model extracts various words and groups of words that are fed into the neural network. Each word is represented by a vector $\bw_i\in \mathbb{R}^{d_w}$. Each group of words is represented by the average of the vectors of each word in the group.
For each mention and pair of mentions, a small number of binary features and distance features are also extracted. Distances and mention lengths are binned into one of the buckets $[0, 1, 2, 3, 4, 5\mathhyphen7, 8\mathhyphen15, 16\mathhyphen31, 32\mathhyphen63, 64+]$ and then encoded in a one-hot vector in addition to being included as continuous features. 
The full set of features is as follows:

\newcommand{\fespace}{\vspace{2.0mm}}
 \vspace{2mm} 
\noindent {\it Embedding Features}: Word embeddings of the head word, dependency parent, first word, last word, two preceding words, and two following words of the mention. Averaged word embeddings of the five preceding words, five following words, all words in the mention, all words in the mention's sentence, and all words in the mention's document.
\fespace 
 
\noindent {\it Additional Mention Features}: The type of the mention (pronoun, nominal, proper, or list), the mention's position (index of the mention divided by the number of mentions in the document), whether the mentions is contained in another mention, and the length of the mention in words.
\fespace 
 
\noindent {\it Document Genre}: The genre of the mention's document (broadcast news, newswire, web data, etc.). 
 
\fespace  
\noindent {\it Distance Features}: The distance between the mentions in sentences, the distance between the mentions in intervening mentions, and whether the mentions overlap. 
\fespace 
 
\noindent {\it Speaker Features}: Whether the mentions have the same speaker and whether one mention is the other mention's speaker as determined by string matching rules from Raghunathan et al.\ (2010)\nocite{raghunathan2010multi}.
\fespace 
 
 \noindent {\it String Matching Features}: Head match, exact string match, and partial string match.
\fespace 

\vspace{1mm}
The vectors for all of these features are concatenated to produce an $I$-dimensional vector $\bh_0$, the input to the neural network. If $a = \text{\sc na}$, the features defined over mention pairs are not included. For this case, we train a separate network with an identical architecture to the pair network except for the input layer to produce anaphoricity scores. 

Our set of hand-engineered features is much smaller than the dozens of complex features typically used in coreference systems. However, we found these features were crucial for getting good model performance. See Section~\ref{sec:mention-ranking-exp} for a feature ablation study. 

\xhdr{Hidden Layers}
The input gets passed through three hidden layers of rectified linear (ReLU) units \cite{nair2010rectified}. Each unit in a hidden layer is fully connected to the previous layer:
\[
	\bh_i(a, m) = \max(0, \bW_i \bh_{i -1}(a, m) + \bb_i)
\]
where $\bW_1$ is a $M_1 \times I$ weight matrix, $\bW_2$ is a $M_2 \times M_{1}$ matrix, and $\bW_3$ is a $d \times M_2$ matrix. 

The output of the last hidden layer is the vector representation for the mention pair: $\repm(a, m) = \bh_3(a, m)$. \\

\subsection{Cluster-Pair Encoder}
\begin{figure}[tb]
\begin{center}
\vspace{0mm}
\includegraphics[width=0.48\textwidth]{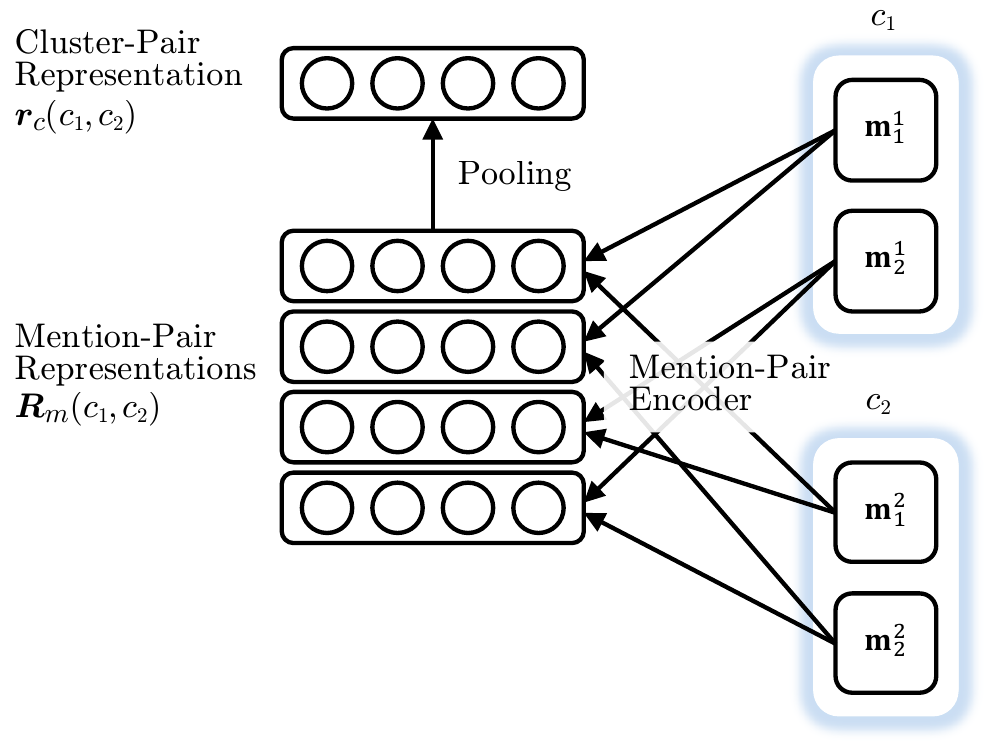}
\vspace{-7mm}
\end{center}
\caption{Cluster-pair encoder.}
\vspace{0mm}
\label{fig:cluster}
\end{figure}

Given two clusters of mentions $c_i = \{m^i_1, m^i_2, ..., m^i_{|c_i|}\}$ and $c_j = \{m^j_1, m^j_2, ..., m^j_{|c_j|}\}$, the cluster-pair encoder produces a distributed representation $\repcl(c_i, c_j) \in \mathbb{R}^{2\dcl}$. The architecture of the encoder is summarized in Figure~\ref{fig:cluster}. 

The cluster-pair encoder first combines the information contained in the matrix of mention-pair representations $\bR_m(c_i, c_j) = [\repm(m^i_1, m^j_1), \repm(m^i_1, m^j_2), ..., \repm(m^i_{|c_i|}, m^j_{|c_j|})]$ to produce $\repcl(c_i, c_j)$. This is done by applying a pooling operation. In particular it concatenates the results of max-pooling and average-pooling, which we found to be slightly more effective than using either one alone:
\[\Resize{0.48\textwidth}{
\repcl(c_i, c_j)_k = 
\begin{cases}
    \text{max} \left\{{\bR_m(c_i, c_j)_{k, \cdot}} \right\}&  \text{for }0 \leq k < d \\
    \text{avg\phantom{.}}\{\bR_m(c_i, c_j)_{k - d, \cdot}\} &  \text{for }d \leq k < 2d 
\end{cases}
}\]

%------------------------------------------------------------------------------------------------

\section{Mention-Ranking Model}
\label{sec:mention-ranking}
Rather than training a cluster-ranking model from scratch, we first train a mention-ranking model that assigns each mention its highest scoring candidate antecedent. There are two key advantages of doing this. First, it serves as pretraining for the cluster-ranking model; in particular the mention-ranking model learns effective weights for the mention-pair encoder. Second, the scores produced by the mention-ranking model are used to provide a measure of which coreference decisions are easy (allowing for an easy-first clustering strategy) and which decisions are clearly wrong (these decisions can be pruned away, significantly reducing the search space of the cluster-ranking model). 

The mention-ranking model assigns a score $\sm(a, m)$ to a mention $m$ and candidate antecedent $a$ representing their compatibility for coreference. This is produced by applying a single fully connected layer of size one to the representation $\repm(a, m)$ produced by the mention-pair encoder:
\[
	\sm(a, m) = \bW_{m} \repm(a, m) + b_{m}
\]
where $\bW_{m}$ is a $1 \times \dm$ weight matrix. At test time, the mention-ranking model links each mention with its highest scoring candidate antecedent. \\

\xhdr{Training Objective} We train the mention-ranking model with the slack-rescaled max-margin training objective from Wiseman et al.\ (2015), which encourages separation between the highest scoring true and false antecedents of the current mention. 
Suppose the training set consists of $N$ mentions $m_1, m_2, ..., m_N$. Let $\mathcal{A}(m_i)$ denote the set of candidate antecedents of a mention $m_i$ (i.e., mentions preceding $m_i$ and \textsc{na}), 
and $\mathcal{T}(m_i)$ denote the set of true antecedents of $m_i$ (i.e., mentions preceding $m_i$ that are coreferent with it or $\{\textsc{na}\}$ if $m_i$ has no antecedent). Let $\hat{t}_i$ be the highest scoring true antecedent of mention $m_i$:
\[
	\hat{t}_i = \underset{t \in \mathcal{T}(m_i)}\argmax \text{\phantom{.}} \sm(t, m_i)
\]
\noindent Then the loss is given by
\[\Resize{0.48\textwidth}{
	\sum\limits_{i=1}^N \underset{a \in \mathcal{A}(m_i)}\max  \Delta(a, m_i)(1 + \sm(a, m_i) - \sm(\hat{t}_i, m_i))
}\]
\noindent where $\Delta(a, m_i)$ is the mistake-specific cost function
\[
    \Delta(a, m_i)= 
\begin{cases}
    \alpha_{\textsc{fn}}& \text{if }a = \textsc{na } \wedge \mathcal{T}(m_i) \neq \{\textsc{na}\}\\
    \alpha_{\textsc{fa}}& \text{if }a \neq \textsc{na } \wedge \mathcal{T}(m_i) = \{\textsc{na}\}\\
    \alpha_{\textsc{wl}}& \text{if }a \neq \textsc{na } \wedge a \notin \mathcal{T}(m_i) \\
    0 & \text{if } a \in \mathcal{T}(m_i)
\end{cases}
\]
for ``false new,'' ``false anaphoric,'' ``wrong link,'' and correct coreference decisions. The different error penalties allow the system to be tuned for coreference evaluation metrics by biasing it towards making more or fewer coreference links. \\

\xhdr{Finding Effective Error Penalties} We fix $\alpha_{\textsc{wl}} = 1.0$ and search for $\alpha_{\textsc{fa}}$ and $\alpha_{\textsc{fn}}$ out of $\{0.1, 0.2, ..., 1.5\}$ with a variant of grid search. Each new trial uses the unexplored set of hyperparameters that has the closest Manhattan distance to the best setting found so far on the dev set. We stopped the search when all immediate neighbors (within 0.1 distance) of the best setting had been explored. 
We found  $(\alpha_{\textsc{fn}}, \alpha_{\textsc{fa}}, \alpha_{\textsc{wl}}) = (0.8, 0.4, 1.0)$ to be best for English and  $(\alpha_{\textsc{fn}}, \alpha_{\textsc{fa}}, \alpha_{\textsc{wl}}) = (0.7, 0.4, 1.0)$ to be best for Chinese on the CoNLL 2012 data. We attribute our smaller false new cost from the one used by Wiseman et al.\ (they set $\alpha_{\textsc{fn}} = 1.2$) to using more precise mention detection, which results in fewer links to {\sc na}. \\

\hyphenation{RMS-prop}
\xhdr{Training Details} We initialized our word embeddings with 50 dimensional ones produced by \texttt{word2vec} \cite{mikolov2013distributed} on the Gigaword corpus for English and 64 dimensional ones provided by Polyglot \cite{al2013polyglot} for Chinese. Averaged word embeddings were held fixed during training while the embeddings used for single words were updated. We set our hidden layer sizes to $M_1 = 1000, M_2 = \dm$ = 500 and minimized the training objective using RMSProp \cite{rmsprop}. To regularize the network, we applied L2 regularization to the model weights and dropout \cite{hinton2012improving} with a rate of 0.5 on the word embeddings and the output of each hidden layer.
 
 \xhdr{Pretraining} As in Wiseman et al.\ (2015)\nocite{wiseman2015learning}, we found that pretraining is crucial for the mention-ranking model's success. We pretrained the network in two stages, minimizing the following objectives from Clark and Manning (2015): \vspace{7pt}  \\
\noindent{\it All-Pairs Classification}
\[\Resize{0.48\textwidth}{
-\sum\limits_{i=1}^N [\sum\limits_{t \in \mathcal{T}(m_i)} \log p(t, m_i) 
+\sum\limits_{f \in \mathcal{F}(m_i)} \log(1 - p(f, m_i)) ]
}\]

\noindent{\it Top-Pairs Classification}
\[\Resize{0.48\textwidth}{
-\sum\limits_{i=1}^N [\max\limits_{t \in \mathcal{T}(m_i)} \log p(t, m_i) 
+\min\limits_{f \in \mathcal{F}(m_i)} \log(1 - p(f, m_i)) ] 
}\]
Where $\mathcal{F}(m_i)$ is the set of false antecedents for $m_i$ and $p(a, m_i) = \text{\it sigmoid}(s(a, m_i))$. The top-pairs objective is a middle ground between the all-pairs classification and mention ranking objectives: it only processes high-scoring mentions, but is probabilistic rather than max-margin. We first pretrained the network with all-pairs classification for 150 epochs and then with top-pairs classification for 50 epochs. See Section~\ref{sec:mention-ranking-exp} for experiments on the two-stage pretraining.

%------------------------------------------------------------------------------------------------

\section{Cluster-Ranking Model}
Although a strong coreference system on its own, the mention-ranking model has the disadvantage of only considering local information between pairs of mentions, so it cannot consolidate information at the entity-level. 
We address this problem by training a cluster-ranking model that scores pairs of clusters instead of pairs of mentions.

Given two clusters of mentions $c_i$ and $c_j$, the cluster-ranking model produces a score $\scl(c_i, c_j)$ representing their compatibility for coreference. This is produced by applying a single fully connected layer of size one to the representation $\repcl(c_i, c_j)$ produced by the cluster-pair encoder:
\[
	\scl(c_i, c_j) = \bW_c \repcl(c_i, c_j) + b_c
\] 
where $\bW_{c}$ is a $1 \times 2\dm$ weight matrix. Our cluster-ranking approach also uses a measure of anaphoricity, or how likely it is for a mention $m$ to have an antecedent. This is defined as
\[
	s_{\textsc{na}}(m) = \bW_{\textsc{na}} \repm(\textsc{na}, m) + b_{\textsc{na}}
\]
where $\bW_{\textsc{na}}$ is a $1 \times \dm$ matrix.

\subsection{Cluster-Ranking Policy Network} 
At test time, the cluster ranker iterates through every mention in the document, merging the current mention's cluster with a preceding one or performing no action. We view this procedure as a sequential decision process where at each step the algorithm observes the current state $x$ and performs some action $u$. 

Specifically, we define a state $x = (C, m)$ to consist of $C = \{c_1, c_2, ...\}$, the set of existing coreference clusters, and $m$, the current mention being considered. At a start state, each cluster in $C$ contains a single mention. Let $\cm \in C$ 
 be the cluster containing $m$ and $\mathcal{A}(m)$ be a set of candidate antecedents for $m$: mentions occurring previously in the document. Then the available actions $U(x)$ from $x$ are
 \vspace{1mm}
\begin{itemize}
\item \textsc{Merge}$[\cm, c]$, where $c$ is a cluster containing a mention in $\mathcal{A}(m)$. This combines $\cm$ and $c$ into a single coreference cluster.
\item \textsc{Pass}. This leaves the clustering unchanged. 
\end{itemize}
 \vspace{0mm}
\noindent After determining the new clustering $C'$ based on the existing clustering $C$ and action $u$, we consider another mention $m'$ to get the next state $x'=(C', m')$.
 
Using the scoring functions $\scl$ and $s_\textsc{na}$, we define a policy network $\pi$ that assigns a probability distribution over $U(x)$ as follows:
\alns{
	\pi(\textsc{Merge}[\cm, c]|x) &\propto e^{\scl(\cm, c)}  \\
	\pi(\textsc{Pass}|x) &\propto e^{s_\textsc{na}(m)}
}
During inference, $\pi$ is executed by taking the highest-scoring (most probable) action at each step. \\

\alglanguage{pseudocode}
\begin{algorithm*}[t]
\caption{Deep Learning to Search}
\begin{algorithmic}
\For {$i=1$ \textbf{to} $num\_epochs$}
	\State Initialize the current training set $\Gamma = \emptyset$
	\For{\textbf{each} example $(x, y) \in \mathcal{D}$}
		\State Run the policy $\pi$ to completion from start state $x$ to obtain a trajectory of states $\{x_1, x_2, ..., x_n\}$
		\For{\textbf{each} state $x_i$ in the trajectory}
			\For{\textbf{each} possible action $u\in U(x_i)$}
				\State Execute $u$ on $x_i$ and then run the reference policy $\pir$ until reaching an end state $e$
				\State Assign $u$ a cost by computing the loss on the end state: $l(u) = \mathcal{L}(e, y)$
			\EndFor
			\State Add the state $x_i$ and associated costs $l$ to $\Gamma$
		\EndFor
	\EndFor
	\State Update $\pi$ with gradient descent, minimizing $\sum\nolimits_{(x, l) \in \Gamma}\sum\nolimits_{u \in U(x)} \pi(u|x)l(u)$
\EndFor
\end{algorithmic}
\label{alg:learn}
\end{algorithm*}

\subsection{Easy-First Cluster Ranking} 
The last detail needed is the ordering in which to consider mentions. Cluster-ranking models in prior work order the mentions according to their positions in the document, processing them left-to-right \cite{rahman2011narrowing,ma2014prune}. However, we instead sort the mentions in descending order by their highest scoring candidate coreference link according to the mention-ranking model. This causes inference to occur in an easy-first fashion where hard decisions are delayed until more information is available. Easy-first orderings have been shown to improve the performance of other incremental coreference strategies \cite{raghunathan2010multi,stoyanov2012easy} because they reduce the problem of errors compounding as the algorithm runs.

We also find it beneficial to prune the set of candidate antecedents $\mathcal A(m)$ for each mention $m$. Rather than using all previously occurring mentions as candidate antecedents, we only include high-scoring ones, which greatly reduces the size of the search space. This allows for much faster learning and inference; we are able to remove over 95\% of candidate actions with no decrease in the model's performance. For both of these two preprocessing steps, we use $s(a, m) - s(\textsc{na}, m)$ as the score of a coreference link between $a$ and $m$.

%------------------------------------------------------------------------------------------------

\subsection{Deep Learning to Search}

We face a sequential prediction problem where future observations (visited states) depend on previous actions. This is challenging because it violates the common i.i.d.\ assumption made in machine learning. Learning-to-search algorithms are effective for this sort of problem, and have been applied successfully to coreference resolution  \cite{daume2005large,clark2015entity} as well as other structured prediction tasks in natural language processing \cite{daume2014efficient,chang2015learningdep}.

We train the cluster-ranking model using a learning-to-search algorithm inspired by SEARN \cite{daume2009search}, which is described in Algorithm~\ref{alg:learn}.  The algorithm takes as input a dataset $\mathcal{D}$ of start states $x$ (in our case documents with each mention in its own singleton coreference cluster) and structured labels $y$ (in our case gold coreference clusters). Its goal is to train the policy $\pi$ so when it executes from $x$, reaching a final state $e$, the resulting loss  $\mathcal{L}(e, y)$ is small. We use the negative of the B$^3$ coreference metric for this loss \cite{bagga1998algorithms}. Although our system evaluation also includes the MUC \cite{vilain1995model} and CEAF{$_{\phi_4}$ \cite{luo2005coreference} metrics, we do not incorporate them into the loss because MUC has the flaw of treating all errors equally and CEAF{$_{\phi_4}$ is slow to compute.

For each example $(x, y) \in \mathcal{D}$, the algorithm obtains a trajectory of states $x_1, x_2, ..., x_n$ visited by the current policy by running it to completion (i.e., repeatedly taking the highest scoring action until reaching an end state) from the start state $x$. This exposes the model to states at train time similar to the ones it will face at test time, allowing it to learn how to cope with mistakes. 

Given a state $x$ in a trajectory, the algorithm then assigns a cost $l(u)$ to each action $u \in U(x)$ by executing the action, ``rolling out" from the resulting state with a reference policy $\pi^\text{ref}$ until reaching an end state $e$, and computing the resulting loss $\mathcal{L}(e, y)$. This rolling out procedure allows the model to learn how a local action will affect the final score, which cannot be otherwise computed because coreference evaluation metrics do not decompose over cluster merges. The policy network is then trained to minimize the risk associated with taking each action: $\sum_{u \in U(x)} \pi(u|x)l(u)$.

Reference policies typically refer to the gold labels to find actions that are likely to be beneficial. Our reference policy $\pi^\text{ref}$ takes the action that increases the B$^3$ score the most each step, breaking ties randomly. It is generally recommended to use a stochastic mixture of the reference policy and the current learned policy during rollouts when the reference policy is not optimal \cite{chang2015learning}. However, we find only using the reference policy (which is close to optimal) to be much more efficient because it does not require neural network computations and is deterministic, which means the costs of actions can be cached. 

\xhdr{Training details} We update $\pi$ using RMSProp and apply dropout with a rate of 0.5 to the input layer. For most experiments, we initialize the mention-pair encoder component of the cluster-ranking model with the learned weights from the mention-ranking model, which we find to greatly improve performance (see Section~\ref{sec:easy}). 

 \xhdr{Runtime} The full cluster-ranking system runs end-to-end in slightly under 1 second per document on the English test set when using a GPU (including scoring all pairs of mentions with the mention-ranking model for search-space pruning). 
 This means the bottleneck for the overall system is the syntactic parsing required for mention detection (about 4 seconds per document on the English test set).

%------------------------------------------------------------------------------------------------

\section{Experiments and Results}
\xhdr{Experimental Setup}
We run experiments on the English and Chinese portions of the CoNLL 2012 Shared Task data \cite{pradhan2012conll}. 
The models are evaluated using three of the most popular coreference metrics: MUC, B$^{3}$, and Entity-based CEAF (CEAF{$_{\phi_4}$). We generally report the average F$_1$ score (CoNLL F$_1$) of the three, which is common practice in coreference evaluation. We used the most recent version of the CoNLL scorer (version 8.01), which implements the original definitions of the metrics. \\

\xhdr{Mention Detection}
Our experiments were run using system-produced predicted mentions. We used the rule-based mention detection algorithm from Raghunathan et al.\ (2010)\nocite{raghunathan2010multi}, which first extracts pronouns and maximal NP projections as candidate mentions and then filters this set with rules that remove spurious mentions such as numeric entities and pleonastic {\it it} pronouns. \\

\subsection{Mention-Ranking Model Experiments}
\label{sec:mention-ranking-exp}
\xhdr{Feature Ablations} We performed a feature ablation study to determine the importance of the hand-engineered features included in our model. The results are shown in Table~\ref{tab:ablation}. We find the small number of non-embedding features substantially improves model performance, especially the distance and string matching features. This is unsurprising, as the additional features are not easily captured by word embeddings and historically such features have been very important in coreference resolvers \cite{bengtson2008understanding}.

\begin{table}[tb]
\begin{tabularx}{\columnwidth}{  X >{\centering\arraybackslash}p{1.7cm} >{\centering\arraybackslash}p{1.8cm} }
  \ttop
  Model & English F$_1$ & Chinese F$_1$ \lone
  Full Model  & 65.52   & 64.41\Tstrut\\ 
 \hspace{3mm}-- {\sc mention} & --1.27   & --0.74\\
 \hspace{3mm}-- {\sc genre} & --0.25   & --2.91\\
 \hspace{3mm}-- {\sc distance} & --2.42   & --2.41\\
 \hspace{3mm}-- {\sc speaker} & --1.26   & --0.93\\
 \hspace{3mm}-- {\sc matching} & --2.07 & --3.44\tbottom 

\end{tabularx}
\precaption
\caption{CoNLL F$_1$ scores of the mention-ranking model on the dev sets without mention, document genre, distance, speaker, and string matching hand-engineered features.}
\vspace{0mm}
\label{tab:ablation}
\end{table}

\xhdr{The Importance of Pretraining} We evaluate the benefit of the two-step pretraining for the mention-ranking model and report results in Table~\ref{tab:pretraining}.
 Consistent with Wiseman et al.\ (2015), we find pretraining to greatly improve the model's accuracy. We note in particular that the model benefits from using both pretraining steps from Section~\ref{sec:mention-ranking}, which more smoothly transitions the model from a mention-pair classification objective that is easy to optimize to a max-margin objective better suited for a ranking task.

\begin{table}[tb]
\tabcolsep=5pt
\begin{tabularx}{\columnwidth}{>{\centering\arraybackslash}p{1.4cm} >{\centering\arraybackslash}p{1.5cm} >{\centering\arraybackslash}p{1.65cm} >{\centering\arraybackslash}p{1.8cm} }
  \ttop
 All-Pairs & Top-Pairs  & English F$_1$ & Chinese F$_1$ \lone
  Yes & Yes   & 65.52 & 64.41 \Tstrut\\
  Yes & No   & --0.36 & --0.24 \\
 No & Yes   & --0.54 & --0.33\\
 No & No   & --3.58 & --5.43\tbottom 
\end{tabularx}
\precaption
\caption{CoNLL F$_1$ scores of the mention-ranking model on the dev sets with different pretraining methods.}
\vspace{0mm}
\label{tab:pretraining}
\end{table} 

 \subsection{Cluster-Ranking Model Experiments}
   \label{sec:easy}
   
 We evaluate the importance of three key details of the cluster ranker: initializing it with the mention-ranking model's weights, using an easy-first ordering of mentions, and using learning to search. The results are shown in Table~\ref{tab:cluster_ablations}. 
 
\begin{table}[tb]
\begin{tabularx}{\columnwidth}{  X >{\centering\arraybackslash}p{1.65cm} >{\centering\arraybackslash}p{1.8cm} } 
  \ttop
 Model           & English F$_1$ & Chinese F$_1$ \lone
  Full Model   & 66.01 & 64.86\Tstrut\\ 
  \hspace{3mm}-- {\sc pretraining}      & --5.01 & --6.85\\
   \hspace{3mm}-- {\sc easy-first}      &--0.15 & --0.12\\
   \hspace{3mm}-- {\sc l2s}      & --0.32 & --0.25\tbottom 
\end{tabularx}
\precaption
\caption{CoNLL F$_1$ scores of the cluster-ranking model on the dev sets with various ablations.\\ -- {\sc pretraining}: initializing model parameters randomly instead of from the mention-ranking model, -- {\sc easy-first}: iterating through mentions in order of occurrence instead of according to their highest scoring candidate coreference link, -- {\sc l2s}: training on a fixed trajectory of correct actions instead of using learning to search.}
\vspace{0mm}
\label{tab:cluster_ablations}
\end{table} 

\xhdr{Pretrained Weights}
We compare initializing the cluster-ranking model randomly with initializing it with the weights learned by the mention-ranking model.
 Using pretrained weights greatly improves performance. We believe the cluster-ranking model has difficulty learning effective weights from scratch due to noise in the signal coming from cluster-level decisions (an overall bad cluster merge may still involve a few correct pairwise links) and the smaller amount of data used to train the cluster-ranking model (many possible actions are pruned away during preprocessing). We believe the score would be even lower without search-space pruning, which stops the model from considering many bad actions.

\xhdr{Easy-First Cluster Ranking}
We compare the effectiveness of easy-first cluster-ranking with the commonly used left-to-right approach. Using a left-to-right strategy simply requires changing the preprocessing step ordering the mentions so mentions are sorted by their position in the document instead of their highest scoring coreference link according to the mention-ranking model. We find the easy-first approach slightly outperforms using a left-to-right ordering of mentions. We believe this is because delaying hard decisions until later reduces the problem of early mistakes causing later decisions to be made incorrectly. 

\xhdr{Learning to Search} We also compare learning to search with the simpler approach of training the model on a trajectory of gold coreference decisions (i.e., training on a fixed cost-sensitive classification dataset). Using this approach significantly decreases performance. We attribute this to the model not learning how to deal with mistakes when it only sees correct decisions during training.
 
\subsection{Capturing Semantic Similarity} Using semantic information to improve coreference accuracy has had mixed in results in previous research, and has been called an ``uphill battle'' in coreference resolution \cite{durrett2013easy}. However, word embeddings are well known for being effective at capturing semantic relatedness, and we show here that neural network coreference models can take advantage of this.

Perhaps the case where semantic similarity is most important is in linking nominals with no head match (e.g., ``the nation'' and ``the country''). We compare the performance of our neural network model with our earlier statistical system \cite{clark2015entity} at classifying mention pairs of this type as being coreferent or not. The neural network shows substantial improvement (18.9 F$_1$ vs. 10.7 F$_1$) on this task compared to the more modest improvement it gets at classifying any pair of mentions as coreferent (68.7 F$_1$ vs. 66.1 F$_1$). Some example wins are shown in Table~\ref{tab:examples}.  
These types of coreference links are quite rare in the CoNLL data (about 1.2\% of the positive coreference links in the test set), so the improvement does not significantly contribute to the final system's score, but it does suggest progress on this difficult type of coreference problem.

\tabcolsep=0.10cm
\begin{table}[t]
\vspace{0mm}
\begin{tabularx}{\columnwidth}{ l l }
  \ttop
  Antecedent          & Anaphor\lone
  the country's leftist rebels & the guerrillas\Tstrut\\
  the company & the New York firm\\
  the suicide bombing & the attack\\
  the gun & the rifle\\
  the U.S. carrier & the ship\tbottom
  \end{tabularx}
     \precaption
 \caption{Examples of nominal coreferences with no head match that the neural model gets correct, but the system from Clark and Manning (2015) gets incorrect.}
 \vspace{0mm}
 \label{tab:examples}
 \end{table}
 
 \begin{table*}[tb]
\small
\setlength{\tabcolsep}{5pt}
\centering
\begin{tabular}{l@{\hskip 13pt} c c c@{\hskip 17pt}   c c c@{\hskip 17pt}   c c c@{\hskip 12pt}   c} 
\ttop
 & \multicolumn{3}{c}{MUC} &  \multicolumn{3}{c}{B$^3$} & \multicolumn{3}{c}{CEAF$_{\phi_4}$} & \Tstrut \\ 
 & Prec. & Rec. & F$_1$ & Prec. & Rec. & F$_1$ & Prec. & Rec. & F$_1$ & Avg. F$_1$ \Bstrut\\ \tline
\multicolumn{11}{c}{\textbf{CoNLL 2012 English Test Data}}  \Tstrut\Bstrut\\ \tline
 Clark and Manning (2015) & 76.12 & 69.38 & 72.59 & 65.64 & 56.01 & 60.44 & 59.44 & 52.98 & 56.02 & 63.02\Tstrut \\ 
 Peng et al.\ (2015)  & -- & -- & 72.22 & -- & -- & 60.50 & -- & -- & 56.37 & 63.03  \\ 
 Wiseman et al.\ (2015) & 76.23 & 69.31 & 72.60 & 66.07 & 55.83 & 60.52 & 59.41 & 54.88 & 57.05 & 63.39\\
 Wiseman et al.\ (2016) & 77.49 & 69.75 & 73.42 & 66.83 & 56.95 & 61.50 & 62.14 & 53.85 & 57.70 & 64.21\Bstrut \\  \mline
  NN Mention Ranker    & 79.77 & 69.10 & 74.05 & 69.68 & 56.37 & 62.32 & 63.02 & 53.59 & 57.92 & 64.76\Tstrut\\
  NN Cluster Ranker     & 78.93 & 69.75 & \textbf{74.06} & 70.08 & 56.98 & \textbf{62.86} & 62.48 & 55.82 & \textbf{58.96} & \textbf{65.29}\tbottom
 \multicolumn{11}{c}{\textbf{CoNLL 2012 Chinese Test Data}}  \Tstrut\Bstrut\\ \tline
 Chen \& Ng (2012)    & 64.69 & 59.92 & 62.21 & 60.26 & 51.76 & 55.69 & 51.61 & 58.84 & 54.99 & 57.63 \Tstrut \\
 Bj{\"o}rkelund \& Kuhn  (2014)  & 69.39 & 62.57 & 65.80 & 61.64 & 53.87 & 57.49 & 59.33 & 54.65 & 56.89 & 60.06  \Bstrut \\  \mline
 NN Mention Ranker   & 72.53 & 65.72 & 68.96 & 65.49 & 56.87 & 60.88 & 61.93 & 57.11 & 59.42 & 63.09\Tstrut\\
 NN Cluster Ranker    & 73.85 & 65.42 & \textbf{69.38} & 67.53 & 56.41 & \textbf{61.47} & 62.84 & 57.62 & \textbf{60.12} & \textbf{63.66}\tbottom
 \end{tabular}
  \precaption
\caption{Comparison with the current state-of-the-art approaches 
 on the CoNLL 2012 test sets. NN Mention Ranker and NN Cluster Ranker are contributions of this work.}
 \vspace{-2mm}
\label{tab:final}
\end{table*} 

\nocite{wiseman2016learning,wiseman2015learning,chen2012combining,bjorkelund2014learning,peng2015joint}

\subsection{Final System Performance}

\begin{figure*}[tb]
\begin{center}
\includegraphics[width=0.9\textwidth]{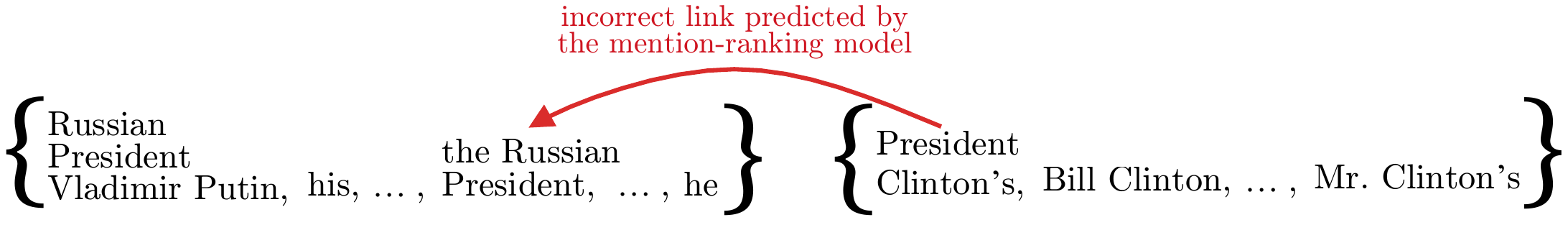}
\vspace{-4mm}
\end{center}
\caption{Thanks to entity-level information, the cluster-ranking model correctly declines to merge these two large clusters when running on the test set. However, the mention-ranking model incorrectly links {\it the Russian President} and {\it President Clinton's}, which greatly reduces the final precision score.}
\label{fig:improvements}
\vspace{1mm}
\end{figure*} 

In Table~\ref{tab:final} we compare the results of our system with state-of-the-art approaches for English and Chinese. Our mention-ranking model surpasses all previous systems.
We attribute its improvement over the neural mention ranker from Wiseman et al.\ (2015) to our model using a deeper neural network, pretrained word embeddings, and more sophisticated pretraining. 

The cluster-ranking model improves results further across both languages and all evaluation metrics, demonstrating the utility of incorporating entity-level information. The improvement is largest in CEAF$_{\phi_4}$, which is encouraging because CEAF$_{\phi_4}$ is the most recently proposed metric, designed to correct flaws in the other two \cite{luo2005coreference}. We believe entity-level information is particularly useful for preventing bad merges between large clusters (see Figure~\ref{fig:improvements}  for an example). 
However, it is worth noting that in practice the much more complicated cluster-ranking model brings only fairly modest gains in performance.

%------------------------------------------------------------------------------------------------

\section{Related Work}
 There has been extensive work on machine learning approaches to coreference resolution \cite{soon2001machine,ng2002improving}, with mention-ranking models being particularly popular \cite{denis2007ranking,durrett2013easy,martschat2015latent}. 
 
 We train a neural mention-ranking model inspired by Wiseman et al.\ (2015) as a starting point, but then use it to pretrain a cluster-ranking model that benefits from entity-level information. Wiseman et al.\ (2016) extend their mention-ranking model by incorporating entity-level information produced by a recurrent neural network running over the candidate antecedent-cluster. However, this is an augmentation to a mention-ranking model, and not fundamentally a clustering model as our cluster ranker is.

Entity-level information has also been incorporated in coreference systems using joint inference \cite{mccallum2003toward,poon2008joint,haghighi2010coreference} and systems that build up coreference clusters incrementally \cite{luo2004mention,yang2008entity,raghunathan2010multi}. We take the latter approach, and in particular combine the cluster-ranking \cite{rahman2011narrowing,ma2014prune} and easy-first \cite{stoyanov2012easy,clark2015entity} clustering strategies. These prior systems all express entity-level information in the form of hand-engineered features and constraints instead of entity-level distributed representations that are learned from data.

We train our system using a learning-to-search algorithm similar to SEARN \cite{daume2009search}. Learning-to-search style algorithms have been employed to train coreference resolvers on trajectories of decisions similar to those that would be seen at test-time by Daum\'e et al.\ (2005), \nocite{daume2005large}Ma et al.\ (2014)\nocite{ma2014prune}, and Clark and Manning (2015). Other works use  structured perceptron models for the same purpose \cite{stoyanov2012easy,fernandes2012latent,bjorkelund2014learning}.

%------------------------------------------------------------------------------------------------

\section{Conclusion} We have presented a coreference system that captures entity-level information with distributed representations of coreference cluster pairs. These learned, dense, high-dimensional feature vectors provide our cluster-ranking coreference model with a strong ability to distinguish beneficial cluster merges from harmful ones. The model is trained with a learning-to-search algorithm that allows it to learn how local decisions will affect the final coreference score. We evaluate our system on the English and Chinese portions of the CoNLL 2012 Shared Task and report a substantial improvement over the current state-of-the-art.

\section*{Acknowledgments}
We thank Will Hamilton, Jon Gauthier, and the anonymous reviewers for their
thoughtful comments and suggestions. This work was supported by NSF Award
IIS-1514268.

%------------------------------------------------------------------------------------------------

\bibliographystyle{acl2016}
\bibliography{clark-manning-acl16-improving}

\end{document}